\documentclass[letterpaper]{article} 

\usepackage{subfigure}
\usepackage{color,xcolor}
\usepackage{amsmath}
\usepackage{amssymb,amsthm}
\usepackage{bm}
\usepackage{multirow}
\usepackage{enumitem}
\usepackage{booktabs}
\usepackage{newfloat}
\usepackage{listings}

\usepackage{aaai24}  
\usepackage{times}  
\usepackage{helvet}  
\usepackage{courier}  
\usepackage[hyphens]{url}  
\usepackage{graphicx} 
\usepackage{natbib}  
\usepackage{caption} 
\usepackage{algorithm}
\usepackage{algorithmic}

\DeclareCaptionStyle{ruled}{labelfont=normalfont,labelsep=colon,strut=off} 
\floatstyle{ruled}
\newfloat{listing}{tb}{lst}{}
\floatname{listing}{Listing}
\title{Towards Automatic Boundary Detection for Human-AI Collaborative Hybrid Essay in Education}
\author{
    Zijie Zeng,
    Lele Sha,
    Yuheng Li,
    Kaixun Yang,
    Dragan Ga\v{s}evi\'{c} \text{and}
    Guanliang Chen\thanks{Corresponding author}
}
\affiliations{
    \textsuperscript{\rm }Centre for Learning Analytics, Monash University, Australia\\
    \{Zijie.Zeng, Lele.Sha1, Yuheng.Li, Kaixun.Yang1, Dragan.Gasevic, Guanliang.Chen\}@monash.edu}

\usepackage{bibentry}

\begin{document}

\maketitle

\begin{abstract}

The recent large language models (LLMs), e.g., ChatGPT, have been able to generate human-like and fluent responses when provided with specific instructions. While admitting the convenience brought by technological advancement, educators also have concerns that students might leverage LLMs to complete their writing assignments and pass them off as their original work. Although many AI content detection studies have been conducted as a result of such concerns, most of these prior studies modeled AI content detection as a classification problem, assuming that a text is either entirely human-written or entirely AI-generated. In this study, we investigated AI content detection in a rarely explored yet realistic setting where the text to be detected is collaboratively written by human and generative LLMs (termed as hybrid text for simplicity). We first formalized the detection task as identifying the transition points between human-written content and AI-generated content from a given hybrid text (boundary detection). We constructed a hybrid essay dataset by partially and randomly removing sentences from the original student-written essays and then instructing ChatGPT to fill in for the incomplete essays. Then we proposed a two-step detection approach where we (1) separated AI-generated content from human-written content during the encoder training process; and (2) calculated the distances between every two adjacent prototypes (a prototype is the mean of a set of consecutive sentences from the hybrid text in the embedding space) and assumed that the boundaries exist between the two adjacent prototypes that have the furthest distance from each other. Through extensive experiments, we observed the following main findings: (1) the proposed approach consistently outperformed the baseline methods across different experiment settings; (2) the encoder training process (i.e., step 1 of the above two-step approach) can significantly boost the performance of the proposed approach; (3) when detecting boundaries for single-boundary hybrid essays, the proposed approach could be enhanced by adopting a relatively large prototype size (i.e., the number of sentences needed to calculate a prototype), leading to a $22$\% improvement (against the best baseline method) in the In-Domain evaluation and an $18$\% improvement in the Out-of-Domain evaluation.

\end{abstract}

\section{Introduction}\label{sec:intro}

The recent advancements in large language models (LLMs) have enabled them to generate human-like and fluent responses when provided with specific instructions. However, the growing generative abilities of LLM have arguably been a sword with two blades. As pointed out by \citet{ma2023abstract,zellers2019defending}, LLMs can potentially be used to generate seemingly correct but unverifiable texts that may be maliciously used to sway public opinion in a variety of ways, e.g., fake news \citep{zellers2019defending}, fake app reviews \citep{martens2019towards}, fake social media posts \citep{fagni2021tweepfake}, and others \citep{weidinger2021ethical, abid2021persistent, gehman2020realtoxicityprompts}. Particularly, concerns have been raised among educators that students may be tempted to leverage the powerful generative capability of LLMs to complete their writing assessments (e.g., essay writing \citep{choi2023chatgpt} and reflective writings \citep{li2023can}), thereby wasting the valuable learning activities that were purposefully designed for developing students' analytical and critical thinking skills \citep{ma2023abstract,dugan2023real,mitchell2023detectgpt}. At the same time, teachers also wasted their effort in grading and providing feedback to artificially generated answers. 
Driven by the above concerns, many studies focused on differentiating human-written content from AI-generated content have been conducted~\citep{mitchell2023detectgpt, ma2023abstract, zellers2019defending, uchendu-etal-2020-authorship, fagni2021tweepfake, ippolito2020automatic}. For example, to advance the techniques for detecting AI-generated content on social media (e.g., Twitter and Facebook), \citet{fagni2021tweepfake} constructed the TweepFake dataset, based on which they evaluated thirteen widely known detection methods. 

\begin{figure}[t]
\centering
\includegraphics[width=8.30cm,height=6.5cm]{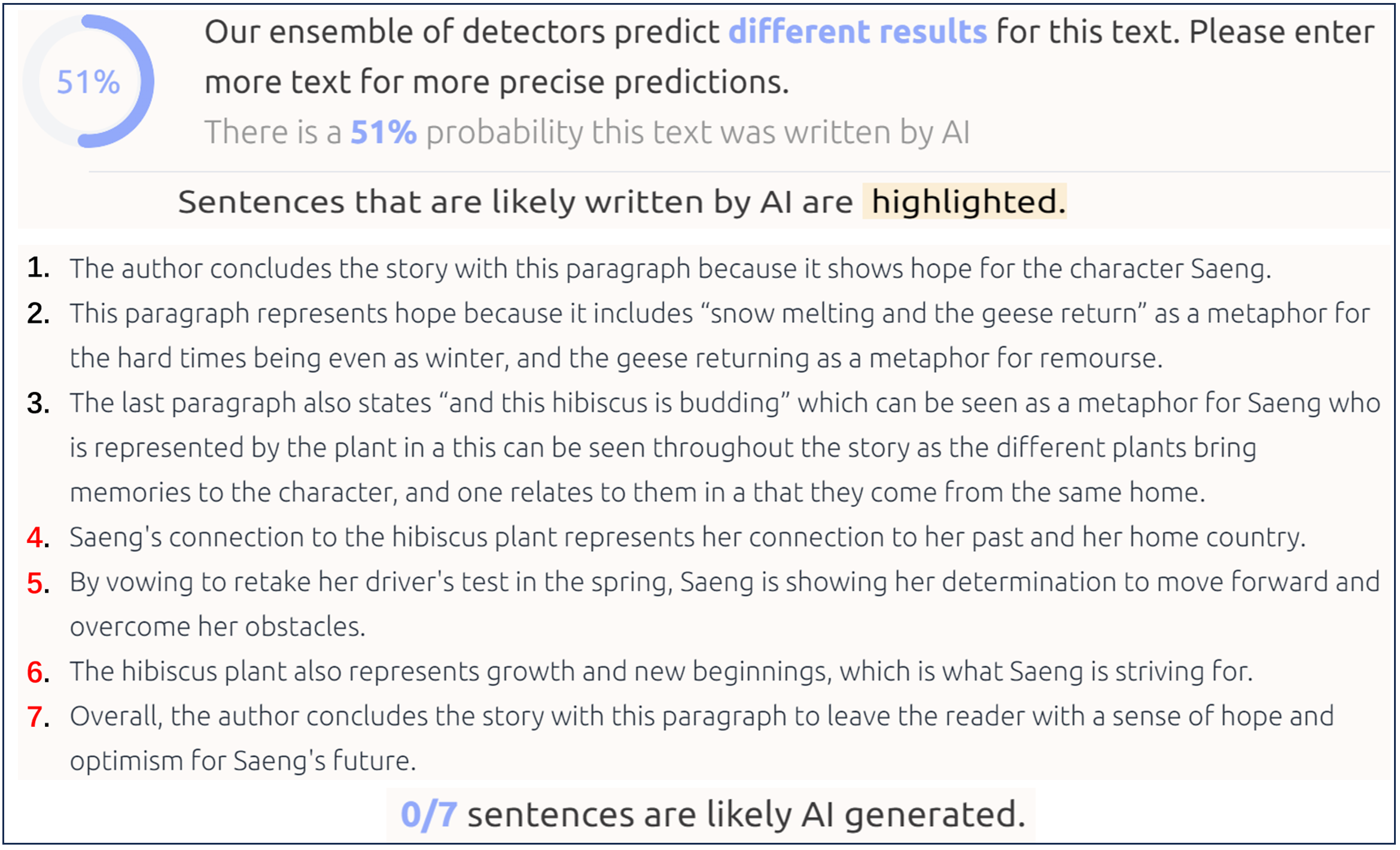}
\caption{An online AI detector GPTZero identified the human-AI collaborative hybrid essay as AI-generated with a 51\% probability but failed to indicate exactly the AI-generated sentences (sentences 4-7).} 

\label{fig1}
\end{figure}

While most existing studies \citep{ma2023abstract, clark2021all, mitchell2023detectgpt, jawahar2020automatic, martens2019towards} formalized the AI-content detection as a binary classification problem and assumed that a sample text is either entirely human-written or entirely AI-generated, \citet{dugan2023real} noticed the trends in human-AI collaborative writing \citep{buschek2021impact,lee2022coauthor} and reflected on the binary classification setting, pointing out that a text (or passage) could begin as human-written and end with AI content generated by LLMs (i.e., hybrid text). They argued that due to the collaborative nature of such text, simply yielding the probability of a text being human-written or AI-generated is less informative than identifying from it the possible AI-generated content. Similar phenomena were observed (illustrated in Figure \ref{fig1}) when we tested a commercial AI content detector called GPTzero\footnote{https://gptzero.me/} with a hybrid essay written first by human (sentences 1-3) and then by ChatGPT (sentences 4-7), the detector suggested that the hybrid input essay was human-written but with mere confidence of 51\% and failed to correctly indicate the AI-generated sentences (sentences 4-7), i.e., none of the sentences were identified as AI-generated. We argue that with the increasing popularity of human-AI collaborative writing, a finer level of AI content detection – specifically, the ability to detect AI-generated portions precisely – will hold greater significance. 
For example, educators might need to know exactly which parts of the text are AI-generated so that they can extract the suspicious parts for further examination (or require them to be revised). Therefore, in this study, we investigated AI content detection in a rarely explored yet realistic setting where the text to be detected is collaboratively written by human and generative LLMs (i.e., hybrid text). As pointed out by \citet{wang2018lstm,wang2021hierarchical}, texts of shorter length (e.g., sentences) are more difficult to classify than longer texts due to the lack of context information (e.g., in Figure \ref{fig1}, GPTZero failed to identify any AI content from the sentence-level). Therefore, instead of addressing AI content detection as a sentence-by-sentence classification problem, we instead followed \citet{dugan2023real} to formalize the AI content detection from hybrid text as identifying the transition points between human-written content and AI-generated content (i.e., boundary detection). We formally defined our research question (RQ) as:
\begin{itemize}
    \item To what extent can the boundaries between human-written content and AI-generated content be automatically detected from essays written collaboratively by students and generative large language models?
\end{itemize}

To investigate this RQ, we constructed a human-AI collaborative hybrid essay dataset by partially and randomly removing sentences from the student-written essays of an open-sourced dataset\footnote{https://www.kaggle.com/c/asap-aes} and then leveraging ChatGPT to fill in for the incomplete essays. Then we proposed our two-step approach (also elaborated in Section \ref{sec:our approach}) to (1) separate AI-generated content from human-written content during the encoder training process; and (2) calculate the distances between every two adjacent prototypes (a prototype \citep{snell2017prototypical} means the average embeddings of a set of consecutive sentences from the hybrid text) and assumed that the boundaries exist between the two adjacent prototypes that have the furthest distance from each other. Through extensive experiments, we summarized the following main findings: (1) the proposed approach consistently outperformed the baseline methods (including a fine-tuned BERT-based method and an online AI detector GPTZero) across the in-domain evaluation and the out-of-domain evaluation; (2) the encoder training process (i.e., step 1 of the above two-step approach) can significantly boost the performance of the proposed approach; (3) when detecting boundaries for single-boundary hybrid essays, the performance of the proposed approach could be enhanced by adopting a relatively large prototype size, resulting in a $22$\% improvement (against the best baseline method) in the In-Domain setting and a $18$\% improvement in the Out-of-Domain setting. Our dataset and the codes are available via Github\footnote{https://github.com/douglashiwo/BoundaryDetection}.

\section{Related Work}\label{sec:background}

\smallskip
\noindent\textbf{Large Language Models and AI Content Detection.}
The recent LLMs have been able to generate fluent and natural-sounding text. Among all LLMs, the Generative Pre-trained Transformer (GPT) \citep{radford2018improving, radford2019language,brown2020language} series of language models developed by OpenAI, are known to be the state of the art and have achieved great success in NLP tasks. ChatGPT, a variant of GPT, is specifically optimized for conversational response generation \citep{hassani2023role} and could generate human-like responses to specific input text (i.e., prompt text) in different styles or tones, according to the role we require it to play \citep{aljanabi2023chatgpt}. 
Along with the advancement of LLMs are concerns over the potential misuse of its powerful generative ability. For example, \citet{ma2023abstract,zellers2019defending} pointed out that LLMs can be used to generate misleading information that might affect public opinion, e.g., fake news \citep{zellers2019defending}, fake app reviews \citep{martens2019towards}, fake social media text \citep{fagni2021tweepfake}, and other harmful text \citep{weidinger2021ethical}. Education practitioners are also concerned about students' leveraging LLMs to complete writing assignments, without developing writing and critical thinking skills \citep{ma2023abstract,dugan2023real,mitchell2023detectgpt}. Driven by the need to identify AI content, many online detecting tools have been developed, e.g., WRITER\footnote{https://writer.com/ai-content-detector/}, Copyleaks\footnote{https://copyleaks.com/ai-content-detector}, and GPTZero. Concurrent with these tools are the AI content detection studies, which have been focused on either (a) investigating humans' abilities to detect AI content \citep{ippolito2020automatic, clark2021all, brown2020language, ethayarajh2022human, dugan2023real}, e.g., \citet{ethayarajh2022human} reported an interesting finding that human annotators rated GPT-3 generated text more human-like than the authentic human-written text; or (b) automating the detection of AI content \citep{ma2023abstract, clark2021all, mitchell2023detectgpt, jawahar2020automatic, martens2019towards}. For example, \citet{ma2023abstract} investigated features for detecting AI-generated scientific text, i.e., writing style, coherence, consistency, and argument logistics. They found that AI-generated scientific text significantly differentiated from human-written scientific text in terms of writing style. 

\smallskip
\noindent\textbf{Human-AI Collaborative Hybrid text and Boundary Detection.}
With modern LLMs, human-AI collaborative writing is becoming more and more convenient. For example, \citet{buschek2021impact} attempted to leverage GPT-2 to generate phrase-level suggestions for human writers in their email writing tasks. Similar studies were conducted in \citet{lee2022coauthor}, where they alternatively used the more powerful GPT-3 for providing sentence-level suggestions for human-AI interactive essay writing. The trends in human-AI collaborative writing also pose a new challenge to the AI content detection research community: \textit{How to detect AI content from a hybrid text collaboratively written by human and LLMs?} As a response to this question, \citet{dugan2023real} proposed to formalize the AI content detection from hybrid text as a boundary detection problem. Specifically, they investigated human's ability to detect the boundary from hybrid texts with one boundary and found that, although humans' abilities to detect boundaries could be boosted after certain training processes, the overall detecting performance was hardly satisfactory, i.e., the human participants could only correctly identify the boundary $23.4\%$ of the time. Similar to \citet{dugan2023real}, our study also targeted boundary detection, but with the following differences: (1) we studied automatic approaches for boundary detection; (2) the hybrid texts considered in this study were of multiple boundaries while \citet{dugan2023real} focused only on single-boundary hybrid texts.

\section{Method}

\subsection{Hybrid Essay Dataset Construction}\label{sec:data}
To the best of our knowledge, there appear to be no available datasets containing hybrid educational texts suitable for investigating our research question (described in Section \ref{sec:intro}). In this section, we set out to construct a hybrid text dataset of educational essays.

\smallskip
\noindent\textbf{Source Data and Pre-processing.}
We identified the essay dataset for the Automated Student Essay Assessment competition\footnote{https://www.kaggle.com/c/asap-aes} as the suitable source material to construct our educational hybrid essay dataset. This source dataset recorded essays from eight question prompts that spanned various topics. These source essays were written by junior high school students (Grade 7 to 10) in the US. To ensure a level of quality and informativeness, we only preserved the source essays with more than 100 words. We noticed that for some source essays, entities such as dates, locations, and the names of the persons had been anonymized and replaced with strings starting with `@' for the sake of privacy protection. For example, the word `Boston' in the source essay was replaced with `@LOCATION'. We filtered out essays containing such entities to prevent our model from incorrectly associating data bias \citep{lyu2023feature} with the label, i.e., associating the presence of `@' with the authorship (human-written or AI-generated) of the text. 

\begin{table*}[!htb]
 \begin{center}

\resizebox{0.94\textwidth}{!}{
     
     \begin{tabular}{l|p{4.7cm}|p{4.7cm}|p{4.7cm}p{4.7cm}} 
    \toprule
     & Task 1 & Task 2 & Task 3 \\ 
    \midrule
    Description & Task 1 requires ChatGPT to generate continuation based on the specified beginning
     text. & Task 2 requires ChatGPT to generate the essay using the specified ending. & Task 3
     requires ChatGPT to fill in between the specified beginning and specified ending. \\ 
    \midrule
    Hybrid Text Structure & H$\rightarrow$G & G$\rightarrow$H & H$\rightarrow$G$\rightarrow$H \\ 
    \midrule
    \#Boundaries & 1 & 1 & 2 \\ 
    \midrule
     & Task 4 & Task 5 & Task 6 \\ 
    \midrule
    Description & Task 4
     requires that the generated essay should include the specified human-written
     text between the beginning and the ending. & Task
     5 requires ChatGPT to fill in for the incompleted essay where some in-between
     text and the ending text have been removed. & Task
     6 requires ChatGPT to fill in for the incompleted essay where some in-between
     text and the beginning text have been removed. \\ 
    \midrule
    Hybrid Text Structure & G$\rightarrow$H$\rightarrow$G & H$\rightarrow$G$\rightarrow$H$\rightarrow$G & G$\rightarrow$H$\rightarrow$G$\rightarrow$H \\ 
    \midrule
    \#Boundaries & 2 & 3 & 3 \\
    \bottomrule
    \end{tabular}

}
 \end{center}
  \caption{Descriptions of the fill-in tasks. H and G in Hybrid Text Structure are short for Human-written text and Generated text, respectively. For example, $<$H$\rightarrow$G$\rightarrow$H$>$ in Task 3 means that the expected hybrid essay should be started/ended with human-written sentences, while the text in between the starting and the ending should be generated by ChatGPT.} \label{table1}
\end{table*}

\smallskip
\noindent\textbf{Hybrid Essay Generation.}
We employed ChatGPT as the generative AI agent for hybrid essay generation, considering its outstanding generative ability \citep{latif2023artificial, xiao2023evaluating} and easy accessibility. To construct a hybrid essay from a source essay $R$, we randomly removed a few sentences\footnote{For a source essay with $k$ sentences, the number of sentences to be removed was randomly selected from $[1, k-1]$.} from $R$ and instructed ChatGPT to perform a fill-in task over the incomplete essay $R'$. Specifically, We designed six fill-in tasks with different prompting texts in order to generate hybrid essays with varying numbers of boundaries, as shown in Table \ref{table1}. 

\smallskip
\noindent\textbf{Prompt Engineering.}
For each hybrid essay, the adopted prompting text generally consisted of two parts: (1) the instructions\footnote{https://www.kaggle.com/competitions/asap-aes/data.} that a writer should refer to when compositing the essay. We directly adopted these instructions as the first part of the prompting text; (2) the second part was the task-relevant prompting text that detailed the specific requirements regarding the structure of the hybrid essay:
\begin{itemize}
    \item Task 1: \textit{Please begin with $<$BEGINNING TEXT$>$.}
    \item Task 2: \textit{Please ensure to use $<$ENDING TEXT$>$ as the ending.}
    \item Task 3: \textit{Please begin with $<$BEGINNING TEXT$>$ and continue writing the second part. For the ending, please use $<$ENDING TEXT$>$ as the ending.}
    \item Task 4: \textit{Please ensure to include $<$IN-BETWEEN TEXT$>$ in between the starting text and the ending text.}
\end{itemize}

Based on the above prompting texts for basic tasks 1--4, we could complete the relatively complex tasks (i.e., Tasks 5 and 6 because multiple missing text pieces needed to be filled to complete these tasks) following two steps:

\begin{itemize}
    \item Step 1: We follow the prompting text of task 3 to generate an initial hybrid essay, which is denoted as $H_1\rightarrow G\rightarrow H_2$, meaning that the first part ($H_1$) and the third part ($H_2$) are \textbf{H}uman-written while the second part ($G$) is \textbf{G}enerated by ChatGPT.
    \item Step 2: We randomly remove\footnote{The number of sentences to be removed from $H_1$ is randomly selected from $[1, k-1]$, where $k$ is the number of sentences of $H_1$.} the first few sentences from $H_1$ and obtain $H_1'$. Then we use the prompt `\textit{Please use $H_1'\rightarrow G\rightarrow H_2$ as the ending}' to obtain the final hybrid essay as required by task 6, which could be denoted as $G'\rightarrow H_1'\rightarrow G\rightarrow H_2$. Similarly, when we remove the last few sentences from $H_2$ and obtain $H_2'$, we can prompt ChatGPT with `\textit{Please begin with $H_1\rightarrow G\rightarrow H_2'$} to obtain the hybrid essay as required by task 5, denoted as $H_1\rightarrow G\rightarrow H_2'\rightarrow G'$.
\end{itemize} 

Due to the randomness\footnote{The extent of the randomness of the ChatGPT-generated output can be controlled by adjusting the hyperparameter `Temperature' (ranging from $[0,1]$), which enables the output to be creative and diverse when set to be high. We used the default value of $0.7$.} of the generative nature of ChatGPT \citep{castro2023large,lyu2023translating}, we could occasionally get invalid\footnote{When the generated output failed to match the expected format as described in Table \ref{fig1}, or contained duplicate sentences, it is considered invalid.} output. To deal with this, we simply discarded the invalid output and instructed ChatGPT to generate a hybrid essay (again) for a maximum of five attempts. If all attempts failed for a specific source essay, we skipped this essay. The statistics of the final hybrid essay dataset are described in Table \ref{table2}.

\begin{table}[!htb]
 \begin{center}

\resizebox{0.44\textwidth}{!}{
     
    \begin{tabular}{l|ccc|c}  
        \toprule
        \multirow{2}{*}{} & \multicolumn{3}{c|}{\#Boundaries} & \multirow{2}{*}{All} \\ 
        \cline{2-4}
         & 1 & 2 & 3 & \\ 
        \midrule
        \#Hybrid essay & 7488 & 6429 & 3219 & 17136 \\ 
        \midrule
        \#Words per essay & 275.3 & 279.5 & 332.6 & 287.6 \\ 
        \midrule
        \#Sentences
         per essay & 12.9 & 13.4 & 16.1 & 13.7 \\ 
        \midrule
        \begin{tabular}[c]{@{}l@{}}Average length\\ of AI-generated\\ sentences\end{tabular} & 22.7 & 21.8 & 21.7 & 22.2 \\ 
        \midrule
        \begin{tabular}[c]{@{}l@{}}Average length\\ of human-written\\ sentences\end{tabular} & 22.7 & 22.6 & 21.2 & 22.4 \\ 
        \midrule
        \begin{tabular}[c]{@{}l@{}}Ratio of AI-generated\\ sentences per essay\end{tabular} & 67.4\% & 58.8\% & 73.2\% & 65.3\% \\
        \bottomrule
    \end{tabular}

}
 \end{center}
 \caption{Statistics of the hybrid essay dataset.} \label{table2}
\end{table}

\subsection{Task and Evaluation}

For a hybrid text $<s_{1}, s_{2},...,s_{n}>$ consisting of $n$ sentences where each sentence is either human-written or AI-generated, the automatic boundary detection task \citep{dugan2023real} requires an algorithm to identify all indexes (boundaries) $i$, where sentence $s_i$ and sentence $s_{i+1}$ are written by different authors, e.g., the former sentence by human and the later sentence by generative language model (e.g., ChatGPT), or vice versa. To evaluate the ability of an algorithm to detect boundaries, we first describe the following two concepts: (1) $L_{topK}$: the list of top-K boundaries suggested by an algorithm; and (2) $L_{Gt}$: the ground-truth list that contains the real boundaries. We adopted the \textit{F1} score as the evaluation metric because it considers both \textit{precision} and \textit{recall} in its calculation (i.e., \textit{F1} is the harmonic mean \citep{lipton2014thresholding} of \textit{precision} and \textit{recall}). The \textit{F1} score can be calculated as:

\begin{equation}\label{eq:1}
    F1@K = 2\cdot  \dfrac{|L_{topK}\cap L_{Gt}|}{ |L_{topK}|+ |L_{Gt}| }.
\end{equation}
Note that $K$ denotes the number of boundary candidates proposed by an algorithm. We set $K=3$ in our study due to the maximum number of boundaries in our dataset being 3. 

\subsection{Boundary Detection Approaches}\label{sec:our approach}

\begin{figure}[t]
    \centering
    \includegraphics[width=8.1cm,height=8.0cm]{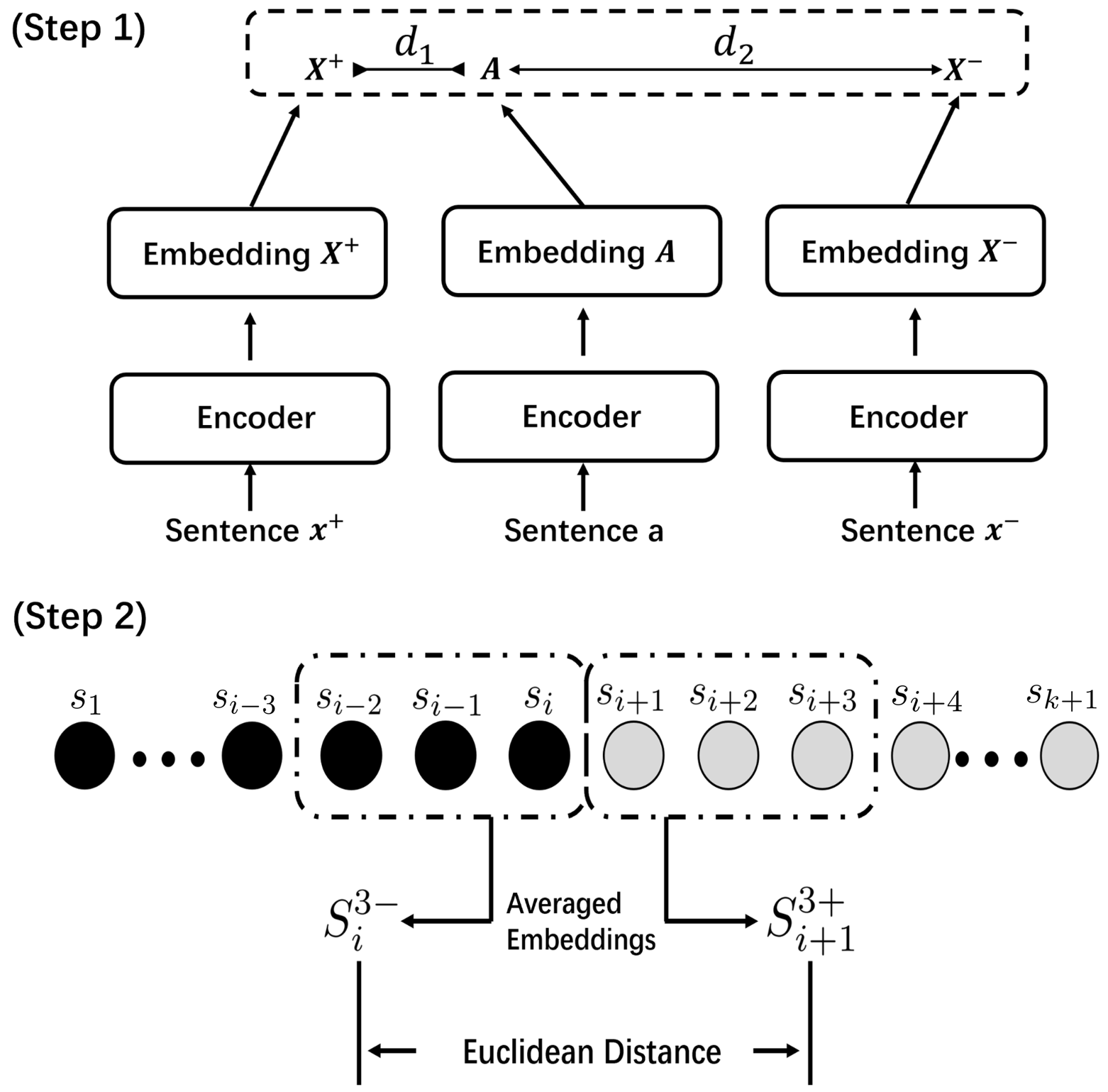}\\
    
    \caption{Our two-step boundary detection approach. The averaged embeddings are also referred to as a prototype. This approach assumes that the boundaries exist between the two adjacent prototypes that have the furthest distance from each other. The prototype size is $p=3$ in this example.} 
    \label{fig2}
\end{figure}

To the best of our knowledge, there are no existing approaches for detecting boundaries from hybrid text with an arbitrary number of sentences and boundaries. Inspired by \citet{perone2018evaluation,liu2020survey}, who pointed out that the quality of representations (embeddings) can significantly affect the performance of downstream tasks, we introduced the following two-step automatic boundary detection approach (see Figure \ref{fig2}):

\begin{itemize}
    \item \textbf{TriBERT (Our two-step approach)}: 
    \begin{enumerate}
        \item We first adopt the pre-trained sentence encoder implemented in the Python package SentenceTransformers\footnote{https://www.sbert.net/} as the initial encoder. We then fine-tune the initial encoder with the triplet BERT-networks \citep{Schroff_2015_CVPR} architecture, which is described as follows: for a sentence triplet $(a, x^+, x^-)$, where $a$ is the anchor sentence whose label is the same as that of sentence $x^+$, but different from the label of sentence $x^-$. The network (i.e., BERT encoder) is trained such that the distance between the embeddings of $a$ and $x^+$ ($d_1$ in step $1$ in Figure \ref{fig2}) is smaller than the distance between the embeddings of $a$ and $x^-$ ($d_2$ in step $1$ in Figure \ref{fig2}). Step 1 aims to separate AI-generated content from human-written content during the encoder training process.
        \item Let $S_i^{p-}$ be the averaged embeddings (also termed as prototype \citep{snell2017prototypical}) of sentence $s_i$ and its $(p-1)$ preceding sentences and $S_{i+1}^{p+}$ be the averaged embeddings of sentence $s_{i+1}$ and its $(p-1)$ following sentences. To identify the possible boundaries, we first calculate the (Euclidean) distances between every two adjacent prototypes $S_i^{p-}$ and $S_{i+1}^{p+}$, where $i\in\{1,2,...,k\}$ and hyperparameter $p$ denotes the number of sentences used to calculate the prototype, i.e., the prototype size. Note that $k+1$ is the number of sentences of the hybrid text. Then we assume that the boundaries exist between the two adjacent prototypes that have the furthest distance from each other. We searched the learning rate from $ \{ 1e-6, 5e-6, 1e-5\}$ and reported the results of prototype size $ p$ in $\{ 1, 2, 3, 4, 5, 6\}$\footnote{Considering an average sentence count of 13.7 per essay (see Table \ref{table2}) in our dataset, we only examined cases where $p \leq 6$.}.

    \end{enumerate}
    
\end{itemize}
We describe all the other baseline approaches as follows:

\begin{itemize}
    
    \item \textbf{BERT}: This method jointly fine-tunes the pre-trained BERT-base encoder and the final dense classification layer. Then the trained model is used to classify each sentence from the hybrid input text. Finally, $i$ is predicted as a boundary if sentence $s_i$ and sentence $s_{i+1}$ are of different predicted labels, e.g., if $s_i$ is predicted as human-written while $s_{i+1}$ is predicted as AI-generated, then this method assumes that $i$ is a possible boundary. This Bert-based classifier was adapted from the pre-trained Bert-based model implemented in the Transformers\footnote{https://github.com/huggingface/transformers} package.
    
    
    \item \textbf{LR}: This method directly employs the sentence encoder implemented in SentenceTransformers for generating sentence embeddings. Then, with the embeddings of each sentence from the hybrid text as input, this method trains a logistic regression binary classifier. Finally, we follow the same process as described in the \textit{BERT} approach to identify the possible boundaries.

    \item \textbf{GPTZero}: We opted for GPTZero due to its API accessibility and its ability to detect on a sentence-by-sentence basis within an input text. This feature facilitates its utilization as a foundational boundary detection benchmark. Specifically, it predicts a label (human-written or AI-generated) for each sentence of the hybrid text. Then, we followed the same process as described in the \textit{BERT} approach to identify the possible boundaries.
    
    \item \textbf{RANDOM}: This method randomly suggests a list of candidate boundaries, serving as a baseline showing the boundary detection performance using random guessing.
\end{itemize}

\section{Experiments}
We conducted both the in-domain evaluation (where models were trained and tested on the same prompts) and the out-of-domain evaluation (where training prompts had no overlap with the testing prompts). Note that our hybrid dataset consists of hybrid essays from eight different prompts.

\subsection{Training, Validating, and Testing}

\smallskip
\noindent\textbf{In-Domain Study.} For the hybrid essays of each prompt, we first grouped the hybrid essays by the source essays from eight prompts. Then we specified that 70\% groups of hybrid essays from each prompt were assigned to the training set. The remaining groups were equally assigned to validation and testing with ratios of 15\% and 15\%, respectively. The process of grouping by source essays was meant to ensure that the source essays for testing would not overlap with the source essays for training, avoiding the situation that essays $E_1$ and $E_2$ were generated from the same source essay $E$ but happened to be assigned to the training set and test set, respectively. Note that in this case, $E_1$ and $E_2$ could theoretically share some common human-written sentences, leading to testing data being exposed in the training stage. 

\smallskip
\noindent\textbf{Out-of-Domain Study.} We followed \citet{jin2018tdnn} to adopt the prompt-wise cross-validation for out-of-domain evaluation, i.e., eight-fold cross-validation in the case of our study because we had hybrid essays from eight different prompts. For each fold, we had the hybrid essays from the target prompt as testing data while hybrid essays from the remaining seven prompts were used for model training. Additionally, 30\% of the training data were held for validation.

\subsection{Parameters, Implementation, and Other Details} For simplicity, we defined the completion of training $n$ samples (in our study we used $n=5000$) as one training epoch and we tested the models on the validation data after each training epoch. Note that the learning rate was reduced by $20\%$ after each epoch and early stopping was triggered at epoch $t$ if the model performance on epoch $t$ showed no improvement as compared to epoch $t-1$. Then, we used the best models (selected based on validation results) to predict on the testing data and reported the results using the F1 metric as described in Equation (\ref{eq:1}). All experiments were run on NVIDIA Tesla T4 GPU with 16 GB RAM.

\section{Results}

We presented the experiment results in Table \ref{table3}, based on which we structured the following analysis and findings.

\begin{table}[!htb]
 \begin{center}

\resizebox{0.45\textwidth}{!}{

\begin{tabular}{l|c|c|c|c} 
\toprule
\multicolumn{5}{c}{\textbf{In-Domain}} \\ 
\midrule
\textbf{Method} & \#\textbf{Bry=1} & \#\textbf{Bry=2} & \#\textbf{Bry=3} & \textbf{All} \\ 
\midrule
\textbf{BERT} & 0.398 & 0.601 & 0.730 & 0.536 \\
\textbf{LR} & 0.265 & 0.377 & 0.404 & 0.332 \\ 
\midrule
\textbf{TriBERT} (p=1) & 0.430 & 0.646 & \textbf{0.752} & 0.571 \\
\textbf{TriBERT} (p=2) & 0.455 & \textbf{0.692} & 0.622 & \textbf{0.575} \\
\textbf{TriBERT} (p=3) & 0.477 & 0.672 & 0.565 & 0.566 \\
\textbf{TriBERT} (p=4) & \textbf{0.486} & 0.641 & 0.514 & 0.549 \\
\textbf{TriBERT} (p=5) & 0.480 & 0.586 & 0.487 & 0.519 \\
\textbf{TriBERT} (p=6) & 0.477 & 0.526 & 0.466 & 0.492 \\ 
\midrule
\multicolumn{5}{c}{\textbf{Out-of-Domain}} \\ 
\midrule
\textbf{Method} & \#\textbf{Bry=1} & \#\textbf{Bry=2} & \#\textbf{Bry=3} & \textbf{All} \\ 
\midrule
\textbf{BERT} & 0.369 & 0.545 & 0.632 & 0.486 \\
\textbf{LR} & 0.159 & 0.248 & 0.241 & 0.208 \\
\textbf{GPTZero} & 0.163 & 0.224 & 0.241 & 0.202 \\ 
\midrule
\textbf{TriBERT} (p=1) & 0.379 & 0.559 & \textbf{0.640} & 0.497 \\
\textbf{TriBERT} (p=2) & 0.417 & \textbf{0.597} & 0.545 & \textbf{0.510} \\
\textbf{TriBERT} (p=3) & 0.421 & 0.567 & 0.463 & 0.484 \\
\textbf{TriBERT} (p=4) & \textbf{0.436} & 0.551 & 0.428 & 0.477 \\
\textbf{TriBERT} (p=5) & 0.424 & 0.511 & 0.402 & 0.452 \\
\textbf{TriBERT} (p=6) & 0.419 & 0.487 & 0.395 & 0.439 \\ 
\midrule
\textbf{TriBERT} (NT,
 p=1) & 0.188 & 0.278 & 0.306 & 0.244 \\
\textbf{TriBERT} (NT,
 p=2) & 0.205 & 0.316 & 0.302 & 0.266 \\
\textbf{TriBERT} (NT,
 p=3) & 0.206 & 0.305 & 0.288 & 0.259 \\
\textbf{TriBERT} (NT,
 p=4) & 0.201 & 0.292 & 0.265 & 0.248 \\
\textbf{TriBERT} (NT,
 p=5) & 0.191 & 0.287 & 0.262 & 0.240 \\
\textbf{TriBERT} (NT,
 p=6) & 0.189 & 0.276 & 0.249 & 0.233 \\ 
\midrule
\textbf{RANDOM} & 0.130 & 0.204 & 0.209 & 0.173 \\
\bottomrule
\end{tabular}

}
 \end{center}
 \caption{Results of different methods on the boundary detection task. We adopted F1 score as the evaluation metric. Note that `NT' in \textit{TriBERT} (NT, $p=k$) means the encoder was used without being fine-tuned. We use \#Bry to denote the number of boundaries. Each reported entry is a mean over three independent runs with the same hyperparameters. The best results are in bold.} \label{table3}
\end{table}

\smallskip
\noindent\textbf{In-Domain (ID) and Out-of-Domain (OOD) Detection. }
We observed that for \textit{LR}, \textit{BERT}, and \textit{TriBERT} ($p=1,2,...$), the ID performance was generally higher than the OOD performance. This observation is not surprising because, in the ID setting, the domains of the test data had all been seen during the training while in the OOD setting, all test domains were unseen during the training stage. Note that for methods involving no training (or fine-tuning) process, i.e., \textit{RANDOM}, \textit{TriBERT} (NT\footnote{`NT' means `Not Trained', i.e., without being fine-tuned.}, $p=1,2,..$) and \textit{GPTZero}, we only reported the OOD performance because all test domains were unseen for these methods.

\smallskip
\noindent\textbf{TriBERT v.s. Baseline Approaches. }
We noticed that \textit{LR} was of similar performance with \textit{GPTZero}, which was better than \textit{RANDOM}, but poorer when compared to \textit{BERT} and \textit{TriBERT} ($p=1,2,...$). Our explanation is that the shallow structure (only one output layer) and the limited number of learnable parameters hampered \textit{LR} from further learning complex concepts from the input sentences. Besides, \textit{TriBERT} ($p=1$) outperformed \textit{BERT} across all levels, i.e., the overall level (\textit{All}) and breakdown levels (\#Bry$=1,2,3$), which demonstrated the advantage of \textit{TriBERT}'s idea of calculating the dissimilarity (Euclidean distance) between every two adjacent prototypes and assuming that the boundaries exist between the most dissimilar (i.e., having the largest distance from each other) adjacent prototypes. 

\smallskip
\noindent\textbf{The Effect of Learning Embeddings through Separating AI-Content from Human-written Content. }
We observed that \textit{TriBERT} (NT, $p=1, 2, ...$), which went through no further encoder training and relied only on the pre-trained encoder from SentenceTransformers for sentence embedding, performed better than \textit{RANDOM}. We also noticed a significant performance improvement from the untrained \textit{TriBERT} (NT, $p=1, 2, ...$) to the fine-tuned \textit{TriBERT} ($p=1, 2, ...$), which demonstrated the necessity of fine-tuning the encoder through separating AI-Content from human-written content before applying the encoder for boundary detection.

\smallskip
\noindent\textbf{The Effect of Varying the Prototype Size of TriBERT. }
To better understand the role of prototype size $p$ in \textit{TriBERT}, we first introduce two effects that can enhance/degrade the prototype in \textit{TriBERT} when varying the prototype size $p$. Let us suppose that we have a set of $k$ consecutive sentences (sharing the same authorship) based on which the prototype will be calculated, denoted as $<s_{i}, s_{i+1},...,s_{i+k-1}>$. Then we would like to introduce a new adjacent sentence $s_{i+k}$ to this sentence pool, i.e., $p$ is growing from $k$ to $k+1$. Note that in this case, the introduction of the new adjacent sentence $s_{i+k}$ can either benefit the prototype (enhancing effect) or degrade the prototype (degrading effect):

\begin{itemize}
    \item \textbf{Enhancing Effect}: If the newly introduced adjacent sentence $s_{i+k}$ shares the same authorship with the existing sentences from the pool, the prototype is enhanced and can yield better representation.
    \item \textbf{Degrading Effect}: If the authorship of $s_{i+k}$ is different from that of the existing sentences, $s_{i+k}$ is considered as noise because the prototype calculated based on hybrid content can represent neither AI content nor human-written content, i.e., the prototype quality is degraded.
\end{itemize}  
From the results of the overall level (i.e., column \textit{All}), we observed that \textit{TriBERT} achieved the best performance with prototype size $p=2$, for which we have the following explanation: when $p<2$, the benefit of increasing $p$ outweighs the risk of bringing noise to the prototype calculation, i.e., the enhancing effect overcomes the degrading effect and plays the dominant role as $p$ grows; however, when $p>=2$, the degrading effect overcomes the enhancing effect and plays the dominant role as $p$ grows, which means \textit{TriBERT}'s performance declines as $p$ grows.

Furthermore, when we dived into the results of the breakdown level (the results of \#Bry $=1,2,3$), we noticed that the best prototype size $p$ tended to be large (or small) if the number of boundaries was small (or large), i.e., the best $p$ for \#Bry $=1,2,3$ were $4$, $2$, and $1$, respectively. Our explanation for this observation is as follows: when we try to sample a set of consecutive sentences $S_k$ (Note that the prototype will be calculated based on $S_k$) from a hybrid text $T$, the more boundaries there are within $T$, the more likely we are to find hybrid content from the selected sentences. For example, consider the hybrid essay $A=<H-H-H-G-G>$ (Here $H$ denotes a human-written sentence and $G$ denotes an AI-generated sentence) with one boundary ($b=1$) located between the third and the fourth sentence. Let $s_i$ and $s_{i+1}$ be two consecutive sentences randomly sampled from $A$ and the probability of $s_i$'s authorship being different from $s_{i+1}$'s is $1/4=25\%$. Similarly, this probability is $4/4=100\%$ for the hybrid essay $B=<H-G-H-G-H>$ sharing the same length with $A$ but with three more boundaries ($b=4$). As a result, \textit{TriBERT} has a higher chance to sample hybrid content (which triggers the degrading effect) for prototype calculation when predicting for hybrid text groups of \#Bry=$2,3$ compared to when predicting for the group with a small boundary number, i.e., \#Bry=$1$. It is also noteworthy that one could alleviate the degrading effect by using a smaller $p$. As can be seen, the best $p$ for the group of \#Bry=$2,3$ is $2$ and $1$, respectively. However, when predicting for the group with a small boundary number (the \#Bry=$1$ group), \textit{TriBERT} is more likely to sample sentences sharing consistent authorship (which triggers the enhancing effect) for prototype calculation, i.e., the prototype is calculated based on purely AI-content (or purely human-written content). In this case, a large prototype size $p$ is preferred for \textit{TriBERT} to improve the detecting performance. As we can see, the best $p$ for \#Bry=$1$ group is when $p=4$, i.e., the proposed \textit{TriBERT} with $p=4$ outperformed the best baseline \textit{BERT} by an improvement of $22$\% in the In-Domain setting and $18\%$ in the Out-of-Domain setting, respectively.

\section{Conclusion and Future Work}\label{sec:conclusion 5}

With the widespread access to generative LLMs (e.g., GPT models), educators are facing unprecedented challenges in moderating the undesirable use of LLMs by students when completing written assessments. Although many prior research efforts have been devoted to the automatic detection of machine-generated text \citep{jawahar2020automatic, clark2021all, mitchell2023detectgpt}, these studies have limited consideration about text data of hybrid nature (i.e., containing both human-written and AI-generated content). To add to the existing studies, we conducted a pioneer investigation into the problem of automatic boundary detection of human-AI hybrid texts in educational scenarios. Specifically, we proposed a two-step boundary detection approach (\textit{TriBERT}) to (1) separate AI-generated content from human-written content during the encoder training process; (2) calculate the (dis)similarity between adjacent prototypes and assume that the boundaries exist between the most dissimilar adjacent prototypes. Our empirical experiments demonstrated that: (1) \textit{TriBERT} outperformed other baseline methods, including a method based on fine-tuned Bert classifier and an online AI detector GPTZero; (2) we further noticed that for hybrid texts with fewer boundaries (e.g., one boundary), \textit{TriBERT} performed well with a large prototype size; When the number of boundaries is large or unclear, a small prototype size is preferred. The above findings can shed light on how to exploit \textit{TriBERT} for better detection performance, e.g., if the hybrid texts are known to be written first by students and then by generative language models (with only one boundary), it will be beneficial to start with a relatively large prototype size. Besides, given the significant advantage of \textit{TriBERT} over the commercial AI detector GPTZero in boundary detection, our \textit{TriBERT} can serve as a supplementary module for AI content detection systems, offering assistance to users who require precise identification of AI-generated content within hybrid text, enabling them to take subsequent actions such as modifying suspicious AI content to reduce its AI-generated appearance, or utilizing the detected text span as preliminary evidence of potential misuse of generative LLMs (or other AI tools). We acknowledge that our hybrid essay generation scheme (Section \ref{sec:data}) is not the only solution to generate hybrid text, e.g., a hybrid text could also be generated collaboratively by humans and generative LLMs through multi-turn interaction \citep{lee2022coauthor}. It is also noteworthy that, boundaries do not exist only BETWEEN sentences, e.g., a boundary can exist WITHIN a sentence that begins as human-written and ends with AI-generated content. As a starter for future work, we would like to investigate boundary detection from hybrid texts generated through multi-turn interaction by students and ChatGPT.

\bibliography{aaai24}

\end{document}